\documentclass[letterpaper, 10 pt, conference]{ieeeconf}

\IEEEoverridecommandlockouts                              
\newif\ifcomment
\commenttrue   

\usepackage{amsmath,mathtools} 
\usepackage{amssymb}  
\usepackage{graphicx}
\usepackage{siunitx}
\usepackage{subcaption}
\usepackage{booktabs}
\usepackage{xcolor}
\usepackage{textcomp}
\usepackage{gensymb} 
\usepackage{multirow} 
\usepackage{comment}
\usepackage{verbatim}
\usepackage[skip=0pt, font=footnotesize]{caption}
\usepackage{bm}
\usepackage[noadjust]{cite}
\usepackage{makecell}
\usepackage{tabularx}
\usepackage{pifont}
\usepackage{layouts}
\usepackage{physics}

\usepackage{amsthm}
\usepackage{algorithm}
\usepackage{algpseudocode}
\usepackage{lipsum}
\usepackage{epstopdf}

\usepackage[inkscapeformat=png]{svg}

\usepackage[english]{babel}
\usepackage[hidelinks]{hyperref}

\usepackage[normalem,normalbf]{ulem}

\newtheoremstyle{colon}%
{}
{}
{\itshape}
{}
{\bfseries}
{:}
{ }
{}
\theoremstyle{colon}

\newtheorem{problem}{Problem}

\theoremstyle{remark}

\makeatletter
\renewcommand{\@IEEEsectpunct}{ \,}
\makeatother

\makeatletter 
\newcommand\Label[1]{&\refstepcounter{equation}(\theequation)\ltx@label{#1}&}
\makeatother

\definecolor{peru}{rgb}{0.803921568627451, 0.5215686274509804, 0.24705882352941178}
\definecolor{violet}{rgb}{0.9333333333333333, 0.5098039215686274, 0.9333333333333333}
\definecolor{greeN}{rgb}{0.17254901960784313, 0.6274509803921569, 0.17254901960784313}
\definecolor{stage0}{RGB}{187,248,255}
\definecolor{stage1}{RGB}{250,255,187}
\definecolor{stage2}{RGB}{187,255,196}
\definecolor{stage0_dark}{RGB}{0,180,200}
\definecolor{stage1_dark}{RGB}{200,180,0}
\definecolor{stage2_dark}{RGB}{0,200,0}

\definecolor{centerline}{RGB}{51,51,255}
\definecolor{exterior}{RGB}{255,153,51}
\definecolor{interior}{RGB}{0,153,0}
\definecolor{slalom}{RGB}{255,51,51}

\def\anonymous{1} 

\newcommand\ringring[1]{%
  {
   \mathop{\kern0pt #1}\limits^{
     \vbox to-1.85ex{
       \kern-2ex 
       \hbox to 0pt{\hss\normalfont\kern.1em \r{}\kern-.45em \r{}\hss}%
       \vss 
     }
   }
  }
}

\renewcommand{\baselinestretch}{0.98}

\newcolumntype{M}[1]{>{\centering\arraybackslash}m{#1}}

\makeatletter
\def\endthebibliography{%
	\def\@noitemerr{\@latex@warning{Empty `thebibliography' environment}}%
	\endlist
}
\makeatother

\IEEEoverridecommandlockouts
\usepackage{tikz}
\usepackage{textcomp}
\usepackage{hyperref}
\usepackage{lipsum}

\newcommand\copyrighttext{%
	\footnotesize Published in IEEE International Conference on Robotics and Automation (ICRA), Yokohama, Japan, 2024.\newline
	 \textcopyright 2024 IEEE. Personal use of this material is permitted.
	Permission from IEEE must be obtained for all other uses, in any current or future media, including reprinting/republishing this material for advertising or promotional purposes, creating new collective works, for resale or redistribution to servers or lists, or reuse of any copyrighted component of this work in other works.
 }
\newcommand\copyrightnotice{%
	\begin{tikzpicture}[remember picture,overlay]
		\node[anchor=south,yshift=10pt] at (current page.south) {\fbox{\parbox{\dimexpr\textwidth-\fboxsep-\fboxrule\relax}{\copyrighttext}}};
	\end{tikzpicture}%
}

\title{\LARGE \bf
Geometric Slosh-Free Tracking for Robotic Manipulators}
\if\anonymous1
\author{Jon Arrizabalaga$^{1}$, Lukas Pries$^{1}$, Riddhiman Laha$^{2}$, Runkang Li$^{2}$, Sami Haddadin$^{2}$, Markus Ryll$^{1,2}$
	\thanks{$^{1}$Autonomous Aerial Systems, School of Engineering and Design,  Technical University of Munich, Germany. E-mail: {\tt\small jon.arrizabalaga@tum.de} and {\tt\small markus.ryll@tum.de}}%
	\thanks{$^{2}$Munich Institute of Robotics and Machine Intelligence (MIRMI), Technical University of Munich}
}
\fi
\begin{document}

\maketitle
\ifcomment
    \copyrightnotice
\fi
\begin{abstract}
    This work focuses on the agile transportation of liquids with robotic manipulators. In contrast to existing methods that are either computationally heavy, system/container specific or dependant on a singularity-prone pendulum model, we present a real-time slosh-free tracking technique. This method solely requires the reference trajectory and the robot's kinematic constraints to output kinematically feasible joint space commands. The crucial element underlying this approach consists on mimicking the end-effector's motion through a virtual quadrotor, which is inherently slosh-free and differentially flat, thereby allowing us to calculate a slosh-free reference orientation. Through the utilization of a cascaded proportional-derivative (PD) controller, this slosh-free reference is transformed into task space acceleration commands, which, following the resolution of a Quadratic Program (QP) based on Resolved Acceleration Control (RAC), are translated into a feasible joint configuration. The validity of the proposed approach is demonstrated by simulated and real-world experiments on a 7 DoF Franka Emika Panda robot.
\end{abstract}
\vspace{-1mm}
\begin{flushleft}
\textbf{Code}: \url{https://github.com/jonarriza96/gsft}\\
\textbf{Video}: \url{https://youtu.be/4kitqYVS9n8}
\end{flushleft}
\vspace{-3.75mm}
\section{Introduction}\label{sec:introduction}
\noindent  Slosh-free liquid handling is of great importance in various fields. For example, it plays a pivotal role in the design of earthquake-resistant watertanks or transportation systems, particularly when handling fuel in planes, propellants in spacecrafts or navigating through waves in ships that have tanks on their decks \cite{abramson1966dynamic}. The use of slosh-free motions is also critical within industrial assembly lines or the healthcare sector, where robotic manipulators are often required to seamlessly mix, transport, or pour liquids \cite{wiesche2003computational}. 

In such manipulation scenarios achieving agile slosh-free maneuverability poses a significant challenge. Humans require exceptional skill to perform such motions effectively. For instance, only the most skilled waiters and waitresses can transport multiple dishes and glasses without spilling any food or liquid. Nevertheless, by utilizing advanced control methods and leveraging the enhanced agility of robotic systems, there exists potential to realize super-human slosh-free motions. Doing so, offers an appealing incentive to the aforementioned applications. Driven by this goal, in this work we focus on the problem of slosh-free tracking with robotic manipulators.

Starting with a purely-scientific interest that led to the formulation of the Navier-Stokes equations \cite{lemarie2018navier}, and further motivated by the economic benefits of multiple industries \cite{wiesche2003computational, abramson1966dynamic}, slosh-free motion control is a well-studied and long-sought problem. In the field of robotics, the existing literature on slosh-free tracking methods with robotic manipulators can be divided into four categories.

The first category relates to the white-box \emph{fluid analysis / modelling} approach, in which the dynamics of the liquid are approximated by models created using complex methods such as Computational Fluid Dynamics (CFD) \cite{muller2003particle, djavareshkian2006simulation} or particle-based approaches \cite{pan2016motion, ichnowski2022gomp}. Once generated, these models can be used for slosh-free trajectory planning and control \cite{kuriyama2008trajectory, pan2016motion}. The computational demands involved in generating such detailed models not only hinder their immediate deployment but also render them case-specific. These models rely on various parameters, including liquid viscosity, density, and vessel size and shape.
\begin{figure}
	\centering
	\captionsetup{type=figure}
	\includegraphics[width=0.87\linewidth]{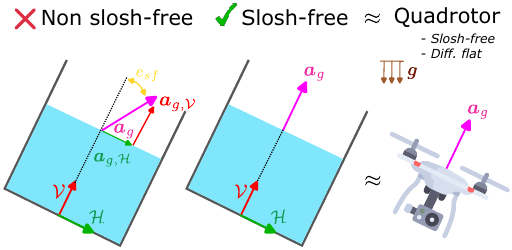}
 \caption{A planar diagram showcasing the slosh-free condition and its equivalence to the quadrotor. Similar to the existing literature our method couples the longitudinal and rotational accelerations in the container to ensure that the resultant translational acceleration acting on the liquid (in magenta) is perpendicular to the liquid's surface. This implies that the slosh-free angle error $e_{sf}$ (in yellow) is desired to be $0$. For this purpose, we emulate the motion of the container with a virtual quadrotor, whose resultant acceleration aligns with the vertical component (in red), and thus, guarantees to be slosh-free.}\label{fig:esf_diagram}
\end{figure}
The second category takes the opposite black-box \emph{learning based} approach, where instead of relying on physics to model the liquid's behavior, the slosh-free motions are Learned from Demonstrations (LfD) \cite{hartz2018adaptive, 6614613}. These methods are also system-specific since they demand extensive training and are exclusively applicable to the system they were trained for.

The third category encompasses the \emph{motion / smoothness minimization} methods. These rely on a first-principles model built from the Navier-Stokes equations, and aim to minimize the slosh-generating oscillations either by using polynomials to shape the input commands \cite{grundelius1999motion, aboel2009design, pridgen2013shaping}, designing smooth velocity profiles \cite{wan2020waiter} or formulating filtering feedback controllers in the frequency domain \cite{yano2001liquid}. Notice that the first two approaches are feedforward, making them vulnerable to unmodeled dynamics and disturbances, whereas the latter requires feedback from the motion of the liquid, thereby creating a dependency on highly specialized sensors designed to measure the state of the liquid. Additionally, these techniques exclusively pertain to translational movements and, as such, are unsuitable for agile motions, as they do not permit tilting of the vessel.

The fourth category overcomes this limitation by coupling the container's translation and rotation by means of a \emph{spherical pendulum model}. These methods cancel out the lateral accelerations acting on the container by emulating the liquid's motion as a virtual pendulum model. Since the original presentation of this technique in \cite{decker2000active,dang2004active}, further studies have allowed for a better understanding on why the augmented pendulum model results in slosh-free motions \cite{han2016study, guang2019dynamic}. 

Out of the extensive literature that falls under the fourth category, two works have shown very appealing results. First, in \cite{biagiotti2018plug} a two-stage plug-in allows to modify the task space commands before being fed to the joint space controller: After passing the desired motion commands through a filter that ensures its smoothness and continuity -- similar to the third category--, a spherical pendulum model is used for calculating the vessel's desired orientation, allowing for compensation of the slosh-generating lateral accelerations. In a similar manner, the authors in the second work \cite{muchacho2022solution} linearize the spherical pendulum model in its lower equilibrium part, resulting in an optimization-based real-time task-space controller for planar (2D) trajectories. Despite these achievements, these two works share three main drawbacks: First, the analytical simplicity of the pendulum's angular parameterization comes at the expense of singularities, preventing it from being applicable to all $\mathrm{SO}(3)$ \cite{lee2017global}. Second, the tracking quality is dependant on the controller design, i.e., the damping-frequency in the first work and the pendulum's rod-length in the second, rendering the overall performance very sensitive to parameter selection. Third, these methods do not tackle the joint space control, leaving it up to inverse kinematics, which is computationally expensive.

This raises the question on how to design a slosh-free tracking method that is (i) singularity-free within all $\mathrm{SE}3$, i.e., spatial (3D) end-effector trajectories in $\mathbb{R}^3$ with a continuous coverage of $\mathrm{SO}3$, (ii) outputs joint commands that fulfill the robot's constraints, (iii) runs online in real-time and (iv) is generic with respect to the liquid properties, cup shape or robot specific parameters. 

To address this question, we present a slosh-free tracking algorithm that, similar to \cite{biagiotti2018plug,muchacho2022solution}, leverages the tilt compensation mechanism. However, instead of relying on the spherical pendulum model, we emulate the end-effector's motion by utilizing the model of a quadrotor. When doing so, we can leverage the differential flatness property of a quadrotor to compute a slosh-free reference pose, i.e. position and orientation. This reference is converted to the joint space by two consecutive controllers, which are not only computationally lightweight but also enforce the robot's kinematic constraints. To the best of the authors' knowledge this is the first method that enables real-time slosh-free tracking of any 3D trajectory with robotic manipulators. 

To achieve this, our method comprises three main ingredients: 1) Using the quadrotor's differential flatness \cite{mellinger2011minimum}, we compute a slosh-free translation and orientation reference. 2) Subsequently, we implement a cascaded proportional derivative (PD) controller to track the desired reference in the task space \cite{aastrom2006advanced}. 3) Last but not least, we utilize Resolved-Acceleration Control (RAC) \cite{luh1980resolved} to formulate a convex Quadratic Program (QP) that maps the desired task space accelerations to joint space while satisfying the robot's kinematic constraints.

\noindent More in detail, we make the following contributions:
\begin{enumerate}
    \item We propose a novel perspective on the problem of slosh-free tracking by identifying the appropriateness of differential-flatness based trajectories from the domain of mobile robotics.  
    \item We formulate an entire pipeline for slosh-free tracking from a desired reference to joint space commands that is (i) capable of tracking any 3D references -- even if infeasible --, (ii) deployable in real-time, (iii) system agnostic and (iv) compliant with the robot's kinematic constraints.
\end{enumerate}

The reminder of this paper is organized as follows: Section~\ref{sec:problem_statement} formally introduces the slosh-free tracking problem tackled in this work and, subsequently, Section~\ref{sec:methodology} presents our solution by delving deeper into all three ingredients. Experimental results are shown in Section~\ref{sec:experiments} before Section~\ref{sec:conclusion} presents the conclusions.

\section{The Slosh-Free Tracking Problem}\label{sec:problem_statement}
\subsection{Robotic manipulator model}
\noindent The forward kinematics of a robotic manipulator are given by a nonlinear mapping between the joint space $\mathcal{J}$ and the task space $\mathcal{T}$, expressed as:
\begin{equation}\label{eq:rm_fk}
    \text{T}_e(t) = 
                \begin{bmatrix}
                    \bm{p}_e(t) & \text{R}_e(t)\\
                    \bm{0} & 1
                \end{bmatrix}= 
                \Upsilon(\bm{q}(t))\,.
\end{equation}
Here, $\mathcal{J}\coloneqq\bm{q}(t)\in\mathcal{Q}\in\mathbb{R}^n$ is the vector of the joint angles, where $n$ is the number of joints, $\mathcal{Q}$ represents the set of kinematically feasible joint configurations and $\mathcal{T}\coloneqq \text{T}_e(t)\in\mathbb{R}^{4x4}\in\mathrm{SE}3$ refers to the homogeneous transformation matrix that represents the end-effector's pose. Its position and orientation are denoted as $\bm{p}_e(t)\in\mathbb{R}^3$ and $\text{R}_e(t)\in\mathrm{SO}3$, respectively. The nonlinear mapping  $\Upsilon(\bm{q}(t))\,:\,\mathbb{R}^n\mapsto\mathrm{SE}3$ can either be computed from Denavit-Hartenberg parameters or from elementary transform sequences \cite{haviland2023dkt1}.

Derivating \eqref{eq:rm_fk} on time allows to compute the end effector's longitudinal and angular velocities $\{\bm{v}_e(t), \bm{\omega}_e(t)\}\in\mathbb{R}^3$ as:
\begin{equation}\label{eq:rm_v}
    \bm{\nu}_e(t)=\left[\bm{v}_e(t),\bm{\omega}_e(t)\right] = J(\bm{q}(t))\dot{\bm{q}}(t)\,,
\end{equation}
where $J(\bm{q}(t))\in\mathbb{R}^{6\times n}$ is the robot's jacobian. Further derivating \eqref{eq:rm_v} on time leads to the end-effector's longitudinal and angular accelerations $\{\bm{a}_e(t), \dot{\bm{\omega}}_e(t)\}\in\mathbb{R}^3$:
\begin{equation}\label{eq:rm_a}
    \bm{\alpha}_e(t)=
    \begin{bmatrix}
        \bm{a}_e(t)\\
        \dot{\bm{\omega}}_e(t)
    \end{bmatrix} = J(\bm{q}(t))\ddot{\bm{q}}(t) + \dot{\bm{q}}(t)\otimes H(\bm{q}(t))\dot{\bm{q}}(t)\,,
\end{equation}
where $H(\bm{q}(t))\in\mathbb{R}^{n\times 6\times n}$ is the robot's hessian and $\otimes$ denotes the tensor contraction.
\subsection{Reference trajectory}
\noindent Let $\Gamma$ refer to a geometric reference whose Cartesian coordinates are given by a time-based reference  $\bm{p}_r\,:\,\mathbb{R}\mapsto\mathbb{R}^3$ that is, at least, $\mathcal{C}^4$:
\begin{equation}\label{eq:geom_ref}
    \Gamma = \{t \in[t_0,t_f] \subseteq\mathbb{R}\mapsto\bm{p}_r(t) \in \mathbb{R}^3\}\,.
\end{equation}
The position difference between the end-effector and the reference will be denoted as \emph{position error} $e_p(t)\in\mathbb{R}$:
\begin{equation}\label{eq:pos_error}
    e_p(t) = ||\bm{p}_r(t) - \bm{p}_e(t)||\,.
\end{equation}

\subsection{Problem statement}
\noindent As initially proposed in \cite{decker2000active,dang2004active} and further discussed in \cite{han2016study,guang2019dynamic}, sloshing can be prevented by suppressing horizontal accelerations acting on the container. This can be achieved by maintaining the surface of the liquid perpendicular to the resultant acceleration at all times. Notice that this resultant acceleration $\bm{a}_g(t)\in\mathbb{R}^3$ includes all accelerations acting on the liquid, and thus the gravity $\bm{g}\in\mathbb{R}^3$ also needs to be considered, i.e., $\bm{a}_g(t) = \bm{a}_e(t) + \bm{g} = \bm{a}_{g,\mathcal{H}}(t) + \bm{a}_{g,\mathcal{V}}(t)$, where $\bm{a}_{g,\mathcal{V}}(t)$ and $\bm{a}_{g,\mathcal{H}}(t)$ refer to the vertical and horizontal components, respectively. In addition, to quantify the slosh-freeness of a given acceleration, we define the \emph{slosh-free error angle}, $e_{sf}\in\mathbb{R}$, as
\begin{equation}\label{eq:sf_angle}
    \tan e_{sf}(t) = \frac{\bm{a}_{g,\mathcal{H}}(t)}{\bm{a}_{g,\mathcal{V}}(t)}\,.
\end{equation}
For a better understanding of these terms, please see the illustrative planar diagram in Fig.~\ref{fig:esf_diagram}.


Having defined the robotic manipulator in \eqref{eq:rm_fk}, the geometric reference $\Gamma$ in \eqref{eq:geom_ref}, as well as the respective position and slosh-free angle errors in eqs.~\eqref{eq:pos_error} and~\eqref{eq:sf_angle}, we can now formally state the problem addressed in this paper:
\begin{problem}[\bfseries Slosh-free tracking for robotic manipulators]\label{problem:sft}
    Given a robotic manipulator whose forward kinematics are expressed in \eqref{eq:rm_fk}, and the geometric reference $\Gamma$ in \eqref{eq:geom_ref}, formulate a control method that achieves:
    \begin{itemize}
        \item[P1.1] \textbf{Reference tracking:} The end-effector tracks the moving reference within a desired tolerance $\epsilon_p$, i.e., $e_p(t)\leq \epsilon_p\,;\forall t\in[t_0,t_f]$.
        \item[P1.2] \textbf{Slosh-free motions:} The tracking motions are conducted in a slosh-free manner, i.e., $|e_{sf}(t)|\leq\epsilon_{sf}\,;\forall t\in[t_0,t_f]$, where $\epsilon_{sf}$ is sufficiently small not to generate any slosh.
        \item[P1.3]  \textbf{Constraint satisfaction:} The kinematic constraints on the robotic manipulator's joints are satisfied for all times, i.e., $\{q(t),\dot{q}(t),\ddot{q}(t), \dddot{q}(t)\}\in\mathcal{Q}\,;\forall t\in[t_0,t_f]$.
    \end{itemize}
\end{problem}

\section{Methodology}\label{sec:methodology}
\noindent The slosh-free tracking scheme presented in this paper leverages (i) a quadrotor differential flatness based reference generator, (ii) a cascaded proportional derivative task space controller and (iii) a joint space controller based on a convex quadratic programme. These three ingredients are the main building blocks of the proposed methodology. In this section, we present further details in each of them. 

\subsection{Quadrotor inspired slosh-free reference generation}\label{sec:df}
\noindent As mentioned earlier, in the last few years, a significant amount of the existing slosh-free control literature has focused on formulating novel architectures that leverage the dynamics of the spherical pendulum model. However, as we have already mentioned in Section~\ref{sec:introduction}, this choice has two main drawbacks: First, the commonly chosen angular parameterization of this model is not applicable to all $\mathrm{SO}(3)$, and second, the pendulum's rod-length introduces a sensitive parameter, leading to a trade-off between tracking quality and slosh-freeness. 

In this work, we take a step back and replace the underlying spherical pendulum model with a virtual quadrotor, a system whose configuration inherently fulfills the slosh-free condition presented in Section~\ref{sec:problem_statement}. The benefits of this change are fourfold: First, the quadrotor's differential flatness allows us to compute slosh-free orientation references for any 3D trajectory, while also being singularity-free and continuous. Second, the dependency on the pendulum's rod-length is dropped. Third, given that the differential flatness mappings are expressed by closed-form non-linear equations, the calculation of a slosh-free reference does not add any computational burden. Fourth, the capacity to map back and forth from the flat outputs to the quadrotor states allows for computing the reference's longitudinal and angular velocity and acceleration, which might render very appealing for the design of task space controllers. 

In the following, we explain how the quadrotor's differential flatness enables the calculation of slosh-free motion references. More specifically, assuming that the end-effector exactly tracks the geometric reference $\Gamma$ in \eqref{eq:geom_ref}, $e_p(t)=0$, while mimicking the motions of a (virtual) quadrotor, we proceed to derive the expressions that facilitate the expansion of the position-reference $\bm{p}_r(t)\in\mathbb{R}^3$ from $\Gamma$ into a full slosh-free pose-reference $\text{T}_{sf,r}=\left[\bm{p}_r(t),\bm{R}_r(t)\right]\in\mathrm{SE}3$. 

The quadrotor's differential flatness property implies that all its states can be written as algebraic functions of four flat outputs $\bm{\sigma}(t) = \left[\bm{p}_r(t), \psi(t) \right]$ and their time derivatives $\dot{\bm{\sigma}}(t),\,\ddot{\bm{\sigma}}(t),\,\dddot{\bm{\sigma}}(t),\,\ddddot{\bm{\sigma}}(t)$. Here $\bm{p}_r(t)$ is the reference's position, which is obliged to emulate the motion of a (virtual) quadrotor, and $\psi(t)$ is the yaw angle, which is not relevant to the slosh-free problem, and thus, will be fixed to a constant value $\tilde{\psi}$. As a consequence, the flat outputs are fully defined by the geometric reference $\Gamma$, i.e., $\bm{\sigma}(t) = \left[\bm{p}_r(t), \tilde{\psi} \right]\equiv\Gamma$. This signifies that all the quadrotor or better named the end-effector states can be computed from reference $\Gamma$. Out of all these states, the successive task space controller solely requires position and orientation slosh-free states $\left[\bm{p}_r(t),\bm{R}_r(t)\right]$, and therefore, we will only show the derivations for these two states. For those interested in the computation of the remaining states, please refer to~\cite{mellinger2011minimum}.

\begin{subequations}\label{eq:df_rot}
Commencing with the position reference $\bm{p}_r(t)$, it becomes evident that it inheretly serves as a flat output that is readily accessible from the reference $\Gamma$. Regarding the orientation reference $\bm{R}_r(t)$, as denoted in the previous paragraph, we parameterize it by a rotation matrix. Its derivation starts by aligning its third component with respect to the resultant acceleration acting on the liquid:
\begin{equation}\label{eq:df_1}
    \bm{z}_r(t) = \frac{\bm{a}_{g}(t)}{||\bm{a}_{g}(t)||}\:\text{with}\:\bm{a}_g = \left[\ddot{\sigma}_1(t),\ddot{\sigma}_2(t),\ddot{\sigma}_3(t)\right] +\bm{g}.
\end{equation}
Next, we define an auxiliary vector 
\begin{equation}\label{eq:df_2}
\tilde{x}_r = \left[\cos{\sigma_4} ,\sin{\sigma_4},0\right]^\intercal\,, 
\end{equation}
that allows us to compute the first and second components of the rotation matrix by
\begin{equation*}
    \bm{y}_r = \frac{\bm{z}_r(t)\times\tilde{\bm{x}}_r}{||\bm{z}_r(t)\times\tilde{\bm{x}}_r||}\,,\quad \bm{x}_r(t) = \bm{y}_r(t)\times\bm{z}_r(t)\,.
\end{equation*}
If $\tilde{\bm{x}}_r(t)\times\bm{z}_r(t)\neq 0$, which can easily be ensured by choosing $\tilde{\psi}$ appropriately, we can stack all three components to finally obtain the reference's slosh-free rotation matrix:
\begin{equation}
    \text{R}_r(t) = \left[\bm{x}_r(t),\bm{y}_r(t),\bm{z}_r(t)\right]\,.
\end{equation}
From eqs.~\eqref{eq:df_1} and~\eqref{eq:df_2} it is apparent that this matrix exclusively depends on the flat outputs and its derivatives. This allows us to compute the reference's slosh-free orientation for the end-effector by simply evaluating the rotation matrix in \eqref{eq:df_rot}.
\end{subequations}

It is worth to highlight that, as per~\eqref{eq:df_1}, the computed slosh-free orientation matrix is dependent on the reference's translational accelerations $\bm{a}_r(t)$. At the same time, the end-effector's orientation is defined by the forward kinematics in \eqref{eq:rm_fk}. Putting both facts together, it results that the reference's acceleration profile $\bm{a}_r(t)$, if tracked by the associated slosh-free orientation $\text{R}_r(t)$ in \eqref{eq:df_rot}, is correlated to the robotic manipulator's joint angles $q(t)$. Intuitively, or following the derivations in~\cite{mellinger2011minimum}, the joint velocities $\dot{q}(t)$ correlate to the jerk profile $\bm{j}_r(t)$ of the reference, while the joint accelerations $\ddot{q}(t)$ correlate to the snap profile $\bm{s}_r(t)$.


\subsection{Task space control: A Cascaded Proportional Derivative}
\noindent The slosh-free pose-reference $\text{T}_{sf,r}=\left[\bm{p}_r(t),\bm{R}_r(t)\right]$ derived in the previous subsection, is now used as the input for a task space controller that outputs longitudinal and angular acceleration commands, i.e., $\bm{u}_\mathcal{T}(t) = \left[\bm{a}_\mathcal{T}(t)\,,\bm{\omega}_\mathcal{T}(t)\right]\in\mathbb{R}^6$. This controller hinges on an error vector which represents the difference between the desired and current poses:
\begin{equation}\label{eq:pose_error}
    \bm{e}_\mathcal{T}(t) = \left[\bm{p}_r(t) - \bm{p}_e(t),\,e_\text{R}(\text{R}_r(t),\text{R}_e(t))\right]\in\mathbb{R}^6\,,
\end{equation}
where $e_\text{R}(\text{R}_r(t),\text{R}_e(t))$ is the error function between the reference and current orientation. To express this function, in this work we adopt the same approach as in \cite{haviland2023dkt1}. Alternatively, one could also consider utilizing the more compact variant proposed in \cite{lee2010geometric}.
\begin{subequations}
As mentioned before, a cascaded PD controller determines the task space acceleration commands. First, given the pose-error $\bm{e}_\mathcal{T}(t)$ defined in~\eqref{eq:pose_error}, the outer loop computes the desired longitudinal and angular end-effector velocities:
\begin{equation}
    \bm{\nu}_\mathcal{T}(t)=\left[\bm{v}_\mathcal{T}(t),\bm{\omega}_\mathcal{T}(t)\right] = \bm{k}_\mathcal{T}\odot \bm{e}_\mathcal{T}(t)\in\mathbb{R}^6\,,
\end{equation}
where $\odot$ refers to the element-wise product and $\bm{k}_\mathcal{T}\in\mathbb{R}^6$ is a proportional gain that controls the convergence rate in the task space. Subsequently, the computed velocities are fed to the inner loop, which, in conjunction with the feedback acquired from the end-effector's longitudinal and angular velocity defined in $\bm{\nu}_e(t)$~\eqref{eq:rm_v}, proceeds to compute  the longitudinal and angular acceleration commands:
\begin{equation}
    \bm{u}_\mathcal{T}(t) =\left[\bm{a}_\mathcal{T}(t)\,,\bm{\omega}_\mathcal{T}(t)\right]= \bm{k}_\nu\odot \left(\bm{\nu}_\mathcal{T}(t)-\bm{\nu}_e(t)\right)\,,
\end{equation}
where $\bm{k}_\mathcal{\nu}\in\mathbb{R}^6$ is also a proportional gain.
\end{subequations}
\begin{figure}
	\centering
	\captionsetup{type=figure}
	\includegraphics[width=\linewidth]{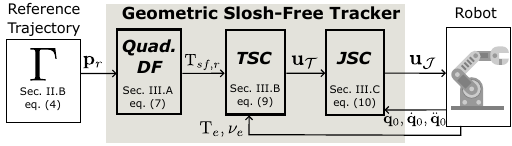}
 \caption{Block diagram of the presented slosh-free tracking method. "Quad. DF." stands for Quadrotor based differential flatness, "TSC" for task space control and "JSC" for joint space control}\label{fig:methodology}
\end{figure}
\begin{figure*}[t]
\centering
\includegraphics[width=\textwidth]{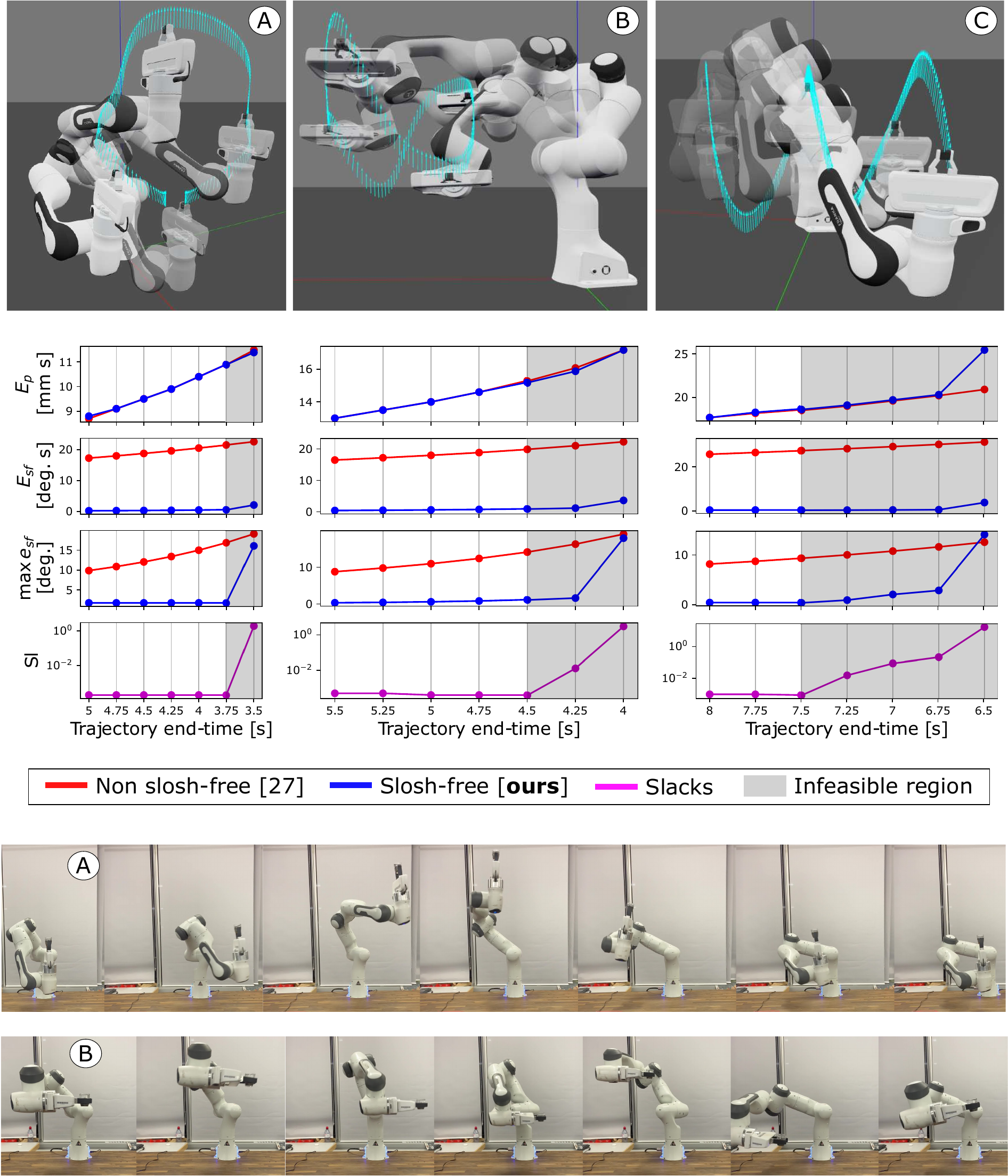}
    \vspace{1mm}
    \caption{\textit{Top (above legend box):} A comparison between the presented geometric slosh-free tracker (blue) against a non-slosh-free standard tracker (red) with a Franka Emika Panda robot for three different case-studies: A) Loop (left col.), B) Lissajous (middle col.), C) Helix (right col.). The motions of the robot along these trajectories are shown in the first row. The cyan arrows result from the differential flatness based reference generation and refer to the required acceleration's direction to ensure the motion to be slosh-free. Each case-study has been evaluated for different trajectory execution times. The respective position errors $E_p$, slosh-free angle errors $E_{sf}$ and maximum slosh-free angle errors $\max{e_{sf}}$ are depicted in the second, third and fourth rows respectively. The slacks resulting from the slosh-free motions are shown in the fifth row. The gray areas refer to the infeasible regions, i.e, the cases where QP~\eqref{eq:QP} would fail to find a solution if slacks would not be activated. \textit{Bottom (below legend box):} The motions (left to right) resulting from running the proposed slosh-free tracker in a real Franka Emika Panda robot for trajectories (A) and (B) with a cup filled of water attached to its end-effector.}\label{fig:simulation_results}
\end{figure*}
\subsection{Joint space control: Optimal Resolved Acceleration}
\noindent The joint space controller maps the task space acceleration commands $\bm{u}_\mathcal{T}(t)$ to the joint space,  $\bm{u}_\mathcal{J}(t)=\left[\bm{q}(t),\dot{\bm{q}}(t),\ddot{\bm{q}}(t)\right]$. For this purpose, we rely on a well-established method denoted as Resolved-Acceleration Control (RAC) \cite{luh1980resolved}, a direct application of the second order differential equation in~\eqref{eq:rm_a}. Unlike the conventional RAC approach, which computes joint accelerations $\ddot{\bm{q}}(t)$ via a nonlinear inversion of ~\eqref{eq:rm_a} and leaves the solution's feasibility unattended, we solve a constrained optimal control problem at each iteration, ensuring the kinematic feasibility of the obtained joint configuration. 

When formulating the optimal control problem, two design choices are particularly noteworthy due to their pivotal roles in enhancing the methodology's robustness and real-time applicability: First, inspired by~\cite{haviland2020purely}, we ensure that the optimal control problem always remains feasible by adding slack variables, $\bm{\delta}\in\mathbb{R}^6$, to the accelerations commanded by the task space controller. These are penalized in the cost function and therefore only get activated if the acceleration commanded by the task space controller is infeasible with respect to the kinematic constraints. Second, to facilitate real-time implementation, we make an approximation by assuming that the Jacobian in~\eqref{eq:rm_v} and the second-order term of~\eqref{eq:rm_a} remain constant. It's worth noting that this assumption holds true when the optimization problem is solved at a high rate, which is the underlying rationale for this approximation. Following this simplification, the optimal control problem becomes a convex Quadratic Programme (QP), and thus, can be very efficiently solved \cite{boyd2004convex}.

\noindent In particular, given an initial joint space configuration $\left[\bm{q}_0,\dot{\bm{q}}_0, \ddot{\bm{q}}_0\right]$, acceleration commands from the task space controller $\bm{u}_\mathcal{\tau}$ and the robot's kinematic constraints, the QP that we solve at every iteration is:
\begin{subequations}\label{eq:QP}
	\begin{alignat}{3}
    \min_{ \bm{q},\dot{\bm{q}}, \ddot{\bm{q}}, \bm{\delta}}& \left[\bm{q},\dot{\bm{q}}, \ddot{\bm{q}}, \bm{\delta}\right] P \left[\bm{q},\dot{\bm{q}}, \ddot{\bm{q}}, \bm{\delta}\right]^\intercal\\   
    \text{s.t.}\quad& \bm{q} = \bm{q}_0 + \dot{\bm{q}}\Delta t\,,\\
	&\bm{\dot{q}} = \bm{\dot{q}}_0 + \ddot{\bm{q}}\Delta t\,,\\
	&\bm{u}_\mathcal{T} + \bm{\delta}=J_0\bm{\ddot{q}} + \dot{\bm{q}}_0\otimes H_0\,\dot{\bm{q}}_0  \,, \label{eq:qp_rac}   \\
    & \left[\,\bm{\uline{q}},\uline{\dot{\bm{q}}}, \uline{\ddot{\bm{q}}}\,\right] \leq \left[\bm{q},\dot{\bm{q}}, \ddot{\bm{q}}\right] \leq  \left[\,\overline{\bm{q}\vphantom{\dot{}}},\overline{\dot{\bm{q}}}, \overline{\ddot{\bm{q}}}\,\right]\,,\label{eq:qp_kin1}\\
	& \uline{\dddot{\bm{q}}}\leq \left(\ddot{\bm{q}} - \ddot{\bm{q}}_0 \right)/\Delta t \leq   \overline{\dddot{\bm{q}}\vphantom{\dot{}}}\,,\label{eq:qp_kin2}
	\end{alignat}
where $P\in\mathbb{R}^{27\times27}$ is a diagonal positive definite matrix, $\Delta t$ is the time-step, $J_0=J(\bm{q_0})$ and $H_0=H(\bm{q_0})$. 
It's worth noting that the resulting QP is convex, with just $27$ variables and all constraints being linear. This illustrates the problem's lightweight nature, and thereby, its real-time applicability.

An overview of the full slosh-free tracking scheme is shown in Fig.~\ref{fig:methodology}, where each block within the highlighted area represents a different component of the proposed slosh-free tracking scheme.
\end{subequations}

\section{Experiments}\label{sec:experiments}
\noindent To assess the validity of our methodology, we conduct simulated and real-world experiments on a 7 DoF Franka Emika Panda robot arm \cite{haddadin2022franka}. 
\subsection{Simulated analysis}
\noindent Aiming to understand the performance of our \emph{slosh-free} tracker, we start by comparing it against a \emph{non slosh-free} variant similar to \cite{haviland2020purely}, where the quadrotor-based differential flatness part is deactivated, and thus, the resulting end-effector's motions are not slosh-free. The comparison is conducted according to four criterion: First, we evaluate the tracking performance by computing the area below position error $E_p=\int_{t_0}^{t_f}\bm{e}_p(t)\,dt\in\mathbb{R}$. Second, we analyze the motion's overall slosh-freeness by computing the area below the slosh-free angle error $E_{sf}=\int_{t_0}^{t_f}e_{sf}(t)\,dt\in\mathbb{R}$. Third, we study the extreme slosh-prone cases by also computing the maximum slosh-free angle error $\max{\bm{e}_{sf}(t)}\in\mathbb{R};\,\forall t\in[t_0,t_f]$. Fourth, we evaluate the feasibility of the motions by computing the sum of the areas below all slack variables $Sl = \sum_{i=0}^6 \int_{t_0}^{t_f} \bm{\delta}_i(t)\,dt \in\mathbb{R}$.

For a better understanding of the solution's performance, we generalize the evaluation by running the comparison in three different case-studies: A) A \emph{loop}-alike trajectory that widely expands the XY plane, B) a \emph{Lissajous} trajectory that requires fast end-effector motions and C) a \emph{helix} that fully exploits the reachable task space. Their differences on shape and scale allow for testing the versatility of the presented method. Moreover, to observe the behavior  of the solution in different dynamic regimes, each case-study is evaluated for different navigation times, i.e., starting with low and feasible durations, up to infeasible ones. This allows us to evaluate the performance of the presented slosh-free tracker in slow-moving, as well as in agile and aggressive maneuvers. 

To conduct the numerical simulations, we use the Robotics Toolbox for Python \cite{rtb}, which provides a convenient interface to the Franka Emika Panda robot. As a solver for the QP in~\eqref{eq:QP} we use quadprog\footnote{quadprog is available in \url{https://github.com/quadprog/quadprog.git}}, an open-source software that efficiently solves convex QPs by relying on the Goldfarb/Idnani dual algorithm \cite{goldfarb1983numerically}. Regarding the selection of gains and weights, in the task space control we heuristically choose $\bm{k}_\mathcal{T}=\left[10,10,10,10,10,10\right]$ and $\bm{k}_\mathcal{\nu} = 10 \bm{k}_\mathcal{T}$. In the joint space control, we define the weights as $P = \text{blkdiag}(1\mathrm{e}{-8}\,I_7,I_7,1\mathrm{e}{-8}\,I_7,1\mathrm{e}{3},I_6)$. All of these values are held constant across all the upcoming evaluations.

The robot motions and the respective benchmarks resulting from running the presented slosh-free tracking method are shown in Fig.~\ref{fig:simulation_results}. The first row depicts the evolution of the robot when navigating along the reference trajectories. The cyan arrows refer to the third component of the slosh-free orientation obtained from the quadrotor-based differential flatness. The second, third and fourth rows depict the aforementioned benchmarking metrics $E_{p},\,E_{sf},\,Sl$ evaluated for different navigation times. The lower the navigation time, the more aggressive the trajectory. The gray areas within these graphs refer to the infeasible region, where the reference's excessive agility implies that if the slacks would not have been included, the accelerations commanded by the task space controller $\bm{u}_\mathcal{T}(t)$ in~\eqref{eq:qp_rac} would be infeasible with respect to the kinematic constraints in~\eqref{eq:qp_kin1} and~\eqref{eq:qp_kin2}, thereby, causing QP~\eqref{eq:QP} to fail. 

The computed motions exhibit three noteworthy characteristics. Firstly, both methods demonstrate equivalent tracking accuracy across all navigation times and case studies. However, the slosh-free variant stands out by delivering a substantial enhancement in the elimination of sloshing. Secondly, in all three cases, the slosh-free variant consistently maintains its slosh-free nature, as indicated by the conditions $e_sf \approx 0$ and $\max{e_sf}<\epsilon_{sf}$. This holds true unless the reference trajectory becomes excessively infeasible. Thirdly, when the navigation time is exceptionally short, implying an overly agile reference trajectory, adjustments in the form of significantly increased slacks are necessary to render the QP~\eqref{eq:QP} feasible. These adjustments, in turn, result in a substantial modification of the accelerations generated by the task space accelerator, ultimately leading to the generation of sloshing in the motion. 

\subsection{Real-world validation}
\noindent Having understood the behavior of the presented slosh-free tracking method, we proceed in performing real-world experiments. For this purpose, we attach a cup of water to the robotic arm's end-effector and test our solution on two of the three case-studies introduced in the simulations: A) the \emph{loop}-alike trajectory evaluated with a navigation time of $3.75$ s and B) the \emph{Lissajous} trajectory with a navigation time of $4.5$ s. The remaining \emph{helix} trajectory is not compatible with the laboratory's specific robot setup\footnote{The \emph{helix} trajectory requires moving the end-effector below its base's height, which would cause a collision with the large table that holds the robot}, and thus, will be dropped. The conducted motions are depicted at the lower part of Fig.~\ref{fig:simulation_results} by image sequences. For a better insight, we encourage readers to view the accompanying video for this paper. As shown by these, and akin to the simulation results, water spillage is successfully avoided.
\section{Conclusion}\label{sec:conclusion}
\noindent In this work we presented a real-time slosh-free tracker for robotic manipulators. Within the proposed framework, the emulation of the end-effector's motion by a virtual quadrotor has allowed for efficiently calculating a slosh-free reference, which, after going through task and joint space controllers, is converted into a feasible joint space configuration. To validate our findings, first, we have conducted a simulation analysis. The results indicate that our solution approximates slosh-free behaviour without compromising tracking performance, even when dealing with infeasible trajectories. Second, we have run some real-world experiments by attaching a cup of water to a Franka Emika Panda robot and showing that the resulting end-effector motions are spill-free.

\newpage
\newcommand{\BIBdecl}{\setlength{\itemsep}{0.18 em}} 
\bibliographystyle{IEEEtran}
\bibliography{Geometric_Slosh_Free_Tracking_for_Robotic_Manipulators}

\end{document}